\documentclass{article}
\usepackage[preprint]{nips_2018}

\usepackage{url,hyperref,lineno,microtype,subcaption}
\usepackage[onehalfspacing]{setspace}
\usepackage{siunitx}
\usepackage{subcaption}

\newcommand{\unit}[2]{\ensuremath{#1\,\mathrm{#2}}}
\usepackage{gensymb}
\usepackage{graphicx}
\usepackage{colortbl}
\usepackage{calc}
\usepackage{multirow}
\usepackage{subcaption}
\usepackage{xcolor}
\usepackage{adjustbox}
\usepackage{hyperref}

\newcommand\setmaximalcolor[1]{\def\maximalcolor{#1}}
\newcommand\setminimalcolor[1]{\def\minimalcolor{#1}}

\usepackage{tikz}
\usetikzlibrary{shapes.arrows,chains}
\usetikzlibrary{positioning}
\usetikzlibrary{shapes.geometric}
\usetikzlibrary{calc}

\definecolor{cred}{RGB}{255,0,0}
\definecolor{cyel}{RGB}{255,234,146}
\definecolor{cgre}{RGB}{ 97,190,130}

\newcommand{\yr}[1]{\cellcolor{cyel!#1!cgre}}
\newcommand{\gy}[1]{\cellcolor{cred!#1!cyel}}

\newcommand{\gyr}[1]{%
    \def\percentagecoloryr{\pgfmathparse{int((#1-\minimalcolor)/(\maximalcolor-\minimalcolor)*200)}}%
    \def\percentagecolorgy{\pgfmathparse{int(2*((#1-\minimalcolor)/(\maximalcolor-\minimalcolor)*100-50))}}%
    \pgfmathparse{ifthenelse((#1-\minimalcolor)/(\maximalcolor-\minimalcolor)>0.5,"1","0")}%
    \let\r\pgfmathresult
    \ifnum\r=0%
        \percentagecoloryr%
        \expandafter\yr\expandafter{\pgfmathresult}%
    \else%
        \percentagecolorgy%
        \expandafter\gy\expandafter{\pgfmathresult}%
    \fi%
    #1%
}

\title{A Differentiable Physics Engine for Deep Learning in Robotics}

\author{%
  Jonas Degrave, Michiel Hermans\thanks{Former member, currently unaffiliated}, Joni Dambre \& Francis wyffels \\
Department of Electronics and Information Systems (ELIS) \\
Ghent University -- iMinds, IDLab \\
Technologiepark-Zwijnaarde 15, B-9052 Ghent, Belgium \\
\texttt{\{Jonas.Degrave,Joni.Dambre,Francis.wyffels\}@UGent.be} \\
}

\begin{document}

\maketitle

\begin{abstract}

An important field in robotics is the optimization of controllers. Currently, robots are often treated as a black box in this optimization process, which is the reason why derivative-free optimization methods such as evolutionary algorithms or reinforcement learning are omnipresent. When gradient-based methods are used, models are kept small or rely on finite difference approximations for the Jacobian. This method quickly grows expensive with increasing numbers of parameters, such as found in deep learning. We propose the implementation of a modern physics engine, which can differentiate control parameters. This engine is implemented for both CPU and GPU. Firstly, this paper shows how such an engine speeds up the optimization process, even for small problems. Furthermore, it explains why this is an alternative approach to deep Q-learning, for using deep learning in robotics. Finally, we argue that this is a big step for deep learning in robotics, as it opens up new possibilities to optimize robots, both in hardware and software. 
\end{abstract}

\section{Introduction}

To solve tasks efficiently, robots require an optimization of their control system. This optimization process can be done in automated testbeds~\citep{degrave2015transfer}, but typically these controllers are optimized in simulation. Standard methods~\citep{aguilar2017stabilization, de2018learning, de2018minimos, meda2018estimation, de2018robust} to optimize these controllers include particle swarms, reinforcement learning, genetic algorithms and evolutionary strategies. These are all derivative-free methods.

A recently popular alternative approach is to use deep Q-learning, a reinforcement learning algorithm. This method requires a lot of evaluations in order to train the many parameters~\citep{levine2016learning}. 
However, deep learning experience has taught us that optimizing with a gradient is often faster and more efficient. This fact is especially true when there are a lot of parameters, as is common in deep learning. However, in the optimization processes for control systems, the robot is almost exclusively treated as a non-differentiable black box. The reason for this is that the robot in hardware is not differentiable, nor are current physics engines able to provide the gradient of the robot models. The resulting need for derivative-free optimization approaches limits both the optimization speed and the number of parameters in the controllers. One could tackle this issue by fitting a neural network model and using its gradient~\citep{grzeszczuk1998neuroanimator}, but those gradients tend to be poor a approximations for the gradient of the original system.

Recent physics engines, such as mujoco~\citep{todorov2012mujoco}, can derive gradients through the model of a robot. However, they can at most evaluate gradients between actions and states in the transitions of the model, and cannot find the derivatives with respect to model parameters.

In this paper, we suggest an alternative approach, by introducing a differentiable physics engine with analytical gradients. This idea is not novel. It has been done before with spring-damper models in 2D and 3D~\citep{hermans2014automated}. This technique is also similar to adjoint optimization, a method widely used in various applications such as thermodynamics~\citep{jarny1991general} and fluid dynamics~\citep{iollo2001aerodynamic}. However, modern engines to model robotics are not based on spring-damper systems. The most commonly used ones are 3D rigid body engines, which rely on impulse-based velocity stepping methods~\citep{erez2015simulation}. In this paper, we test whether these engines are also differentiable and whether this gradient is computationally tractable. We will show how this method does speed up the optimization process tremendously, and give some examples where we optimize deep learned neural network controllers with millions of parameters.

\section{Materials and methods}

\subsection{A 3D Rigid Body Engine}

The goal is to implement a modern 3D rigid body engine, in which parameters can be differentiated with respect to the fitness a robot achieves in a simulation, such that these parameters can be optimized with methods based on gradient descent. 

The most frequently used simulation tools for model-based robotics, such as PhysX, Bullet, Havok and ODE, go back to MathEngine~\citep{erez2015simulation}. These tools are all 3D rigid body engines, where bodies have 6 degrees of freedom, and the relations between them are defined as constraints. These bodies exert impulses on each other, but their positions are constrained, e.g. to prevent the bodies from penetrating each other. The velocities, positions and constraints of the rigid bodies define a linear complementarity problem (LCP)~\citep{chappuis2013constraints}, which is then solved using a Gauss-Seidel projection (GSP) method~\citep{jourdan1998gauss}. The solution of this problem are the new velocities of the bodies, which are then integrated by semi-implicit Euler integration to get the new positions~\citep{stewart2000implicit}. This system is not always numerically stable. Therefore the constraints are usually softened~\citep{catto2009modeling}.

The recent growth of automatic differentiation libraries, such as Theano~\citep{team2016theano}, Caffe~\citep{jia2014caffe} and Tensorflow~\citep{tensorflow2015-whitepaper}, has allowed for efficient differentiation of remarkably complex functions before~\citep{degrave2016spatial}. Therefore, we implemented such a physics engine from scratch as a mathematical expression in Theano, a software library which does automatic evaluation and differentiation of expressions with a focus on deep learning.  The resulting computational graph to evaluate this expression is then compiled for both CPU and GPU. To be able to compile for GPU however, we had to limit our implementation to a restricted set of elementary operations. The range of implementable functions is therefore severely capped. However, since the analytic gradient is determined automatically, the complexity of correctly implementing the differentiation is removed entirely.

One of these limitations with this restricted set of operations, is the limited support for conditionals. Therefore we needed to implement our physics engine without branching, as this is not yet available in Theano for GPU. Note that newer systems for automatic differentiation such as PyTorch~\cite{paszke2017automatic} do allow branching. Therefore we made sacrificed some abilities of our system. For instance, our system only allows for contact constraints between different spheres or between spheres and the ground plane. Collision detection algorithms for cubes typically have a lot of branching~\citep{mirtich1998v}. However, this sphere based approach can in principle be extended to any other shape~\citep{hubbard1996approximating}. On the other hand, we did implement a rather accurate model of servo motors, with gain, maximal torque, and maximal velocity parameters.

Another design choice was to use rotation matrices rather than the more common quaternions for representing rotations. Consequently, the states of the bodies are larger, but the operations required are matrix multiplications. This design reduced the complexity of the graph. However, cumulative operations on a rotation matrix might move the rotation matrix away from orthogonality. To correct for this, we renormalize our matrix with the update equation~\citep{premerlani2009direction}:

\begin{equation}
A' = \frac{3A - A \circ (A \cdot A)}{2}
\end{equation}

where $A'$ is the renormalized version of the rotation matrix $A$. `$\circ$' denotes the elementwise multiplication, and `$\cdot$' the matrix multiplication.

These design decisions are the most important aspects of difference with the frequently used simulation tools. In the following section, we will evaluate our physics simulator on some different problems. We take a look at the speed of computation and the number of evaluations required before the parameters of are optimized.

\subsubsection{Throwing a Ball}

To test our engine, we implemented the model of a giant soccer ball in the physics engine, as shown in Figure~\ref{fig:3}\textbf{(A)}. The ball has a \unit{1}{m} diameter, a friction of $\mu=1.0$ and restitution $e=0.5$. The ball starts off at position $(0,0)$. After \unit{5}{s} it should be at position $(10,0)$ with zero velocity $v$ and zero angular velocity $\omega$. We optimized the initial velocity $v_0$ and angular velocity $\omega_0$ at time $t=\unit{0}{s}$ until the errors at $t=\unit{5}{s}$ are less than \unit{0.01}{m} and \unit{0.01}{m/s} respectively.

Since the quantity we optimize is only know at the end of the simulation, but we need to optimize the parameters at the beginning of the simulation, we need to backpropagate our error through time~(BPTT)~\citep{sutskever2013training}. This approach is similar to the backpropagation through time method used for optimizing recurrent neural networks~(RNN). In our case, every time step in the simulation can be seen as one pass through a neural network, which transforms the inputs from this timestep to inputs for the next time step. For finding the gradient, this RNN is unfolded completely, and the gradient can be obtained by differentiating this unfolded structure. This analytic differentiation is done automatically by the Theano library.

Optimizing the six parameters in $v_0$ and $\omega_0$ took only 88~iterations with gradient descent and backpropagation through time. Optimizing this problem with CMA-ES~\citep{hansen2006cma}, a state of the art derivative-free optimization method, took 2422~iterations. Even when taking the time to compute the gradient into account, the optimization with gradient descent takes \unit{16.3}{s}, compared to \unit{59.9}{s} with CMA-ES. This result shows that gradient-based optimization of kinematic systems can in some cases already outperform gradient-free optimization algorithms from as little as six parameters.

\subsection{Policy Search}

\tikzstyle{body}=[black, fill=green]
\tikzstyle{legpart}=[ultra thick]
\tikzstyle{legjoint}=[draw,circle,minimum size=0.1em,fill=green!40]
\tikzstyle{line}=[-,thick]
\tikzstyle{arrow}=[->,thick]
\tikzstyle{verticalin}=[in=90]
\tikzstyle{horizontalin}=[in=180]
\tikzstyle{block} = [draw,fill=blue!15,minimum size=2em]
\tikzstyle{var} = [draw,circle,fill=blue!15,minimum size=2em]

\tikzset{
    pic shift/.store in=\shiftcoord,
    pic shift={(0,0)},
    robotsymbol/.pic={
        \begin{scope}[shift={\shiftcoord}]
            \path (-0.0,0.05) coordinate (-body);
            \draw[body] (-0.55,0.05) rectangle (0.55,-0.15);
            
            \path[legpart, orange] (-0.5,-0.05) edge (-0.4,-0.42);
            \path[legpart, red] (-0.4,-0.42) edge (-0.6,-0.7);
            
            \path[legpart, orange] (0.5,-0.05) edge (0.6,-0.42);;
            \path[legpart, red] (0.6,-0.42) edge (0.4,-0.7);
            
            \path (0.0, -0.7) coordinate (-norm);
            \draw[line,verticalin,out=0] (-0.5,-0.05) to (-norm);
            \draw[line,verticalin,out=0] (-0.4,-0.42) to (-norm);
            \draw[line,verticalin,out=180] (0.5,-0.05) to (-norm);
            \draw[line,verticalin,out=180] (0.6,-0.42) to (-norm);        
        \end{scope}
    }
}
    
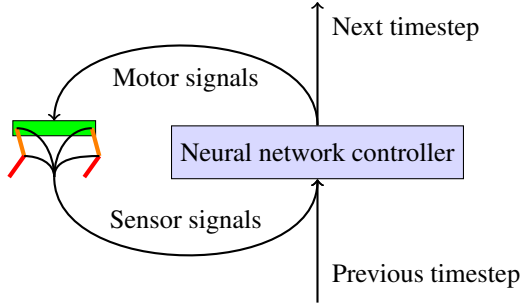
\begin{figure}[tpb]
\begin{center}

\begin{tikzpicture}
    \path (0.0, 0.0) coordinate (body);    
    \pic (rob) at (0,0) {robotsymbol};
    \node[block] (nn) at (3.5, -0.37) {Neural network controller};

    \node[rotate=0,anchor=center] at (1.75,-1.3) {Sensor signals};
    \node[rotate=0,anchor=center] at (1.75, 0.6) {Motor signals};
    \draw[arrow,in=-90,out=-90] (rob-norm) to (nn);
    \draw[arrow,in=90,out=90] (nn) to (rob-body);
    \path (nn)+(0.0,2.0) coordinate (next);
    \path (nn)+(-0.0,-2.0) coordinate (prev);
    \draw[arrow] (nn) to (next);
    \draw[arrow] (prev) to (nn);
    \node[anchor=west, below right=0.1 of next] (next_t) {Next timestep};
    \node[anchor=west, above right=0.1 of prev] (prev_t) {Previous timestep};
\end{tikzpicture}
    
\end{center}
\caption{Illustration of how a closed loop neural network controller would be used to actuate a robot. The neural network receives sensor signals from the sensors on the robot and uses these to generate motor signals which are sent to the servo motors. The neural network can also generate a signal which it can use at the next timestep to control the robot.} \label{fig:1}
\end{figure}

To evaluate the relevance of our differentiable physics engine, we use a neural network as a general controller for a robot, as shown in Figure~\ref{fig:1}. We consider a general robot model in a discrete-time dynamical system $\mathbf{x}^{t+1} = f_{\textrm{ph}}(\mathbf{x}^t, \mathbf{u}^t)$ with a task cost function of $l(\mathbf{x}^t, \mathbf{p})$, where $\mathbf{x}^t$ is the state of the system at time $t$ and $\mathbf{u}^t$ is the input of the system at time $t$. $\mathbf{p}$ provides some freedom in parameterizing the loss. If $X^t$ is the trajectory of the state up to time $t-1$, the goal is to find a policy $u^t = \pi(X^t)$ such that we minimize the loss $\mathcal{L}_\pi$.

\begin{equation}
\label{eq:L}
\mathcal{L}_\pi = \sum\limits_{t=0}^{T} l(\mathbf{x}^t, \mathbf{p}) \\
\textnormal{s.t.} \quad \mathbf{x}^{t+1} = f_{\textrm{ph}}(\mathbf{x}^t, \pi(X^t)) \quad \textnormal{and} \quad \mathbf{x}^0 = x^{\textrm{init}}
\end{equation}

In previous research, finding a gradient for this objective has been described as presenting challenges~\citep{mordatch2014combining}. An approximation to tackle these issues has been discussed in~\citet{levine2013variational}.

We implement this equation into an automatic differentiation library, ignoring these challenges in finding the analytic gradient altogether. The automatic differentiation library, Theano in our case, analytically derives this equation and compiles code to evaluate both the equation and its gradient.

Unlike in previous approaches such as iLQR~\cite{todorov2005generalized} and DDP~\cite{bertsekas2005dynamic}, we propose not to use this gradient to optimize a trajectory, but to use the gradient obtained to optimize a general controller parameterized by a neural network. This limits the amount of computation at execution time, but requires the optimization of a harder problem with more parameters.

We define our controller as a deep neural network $g_{\textrm{deep}}$ with weights $\mathbf{W}$. We do not pass all information $X^t$ to this neural network, but only a vector of values $\mathbf{s}^t$ observed by the modeled sensors $s(\mathbf{x}^t)$. We also provide our network with (some of the) task-specific parameters $\mathbf{p}'$. Finally, we add a recurrent connection to the controller in the previous timestep $\mathbf{h}^t$.  Therefore, our policy is the following:
\begin{equation}
\label{eq:g}
\pi(X^t)=g_{\textrm{deep}}(s(\mathbf{x}^t), \mathbf{h}^t, \mathbf{p}' \; | \;  \mathbf{W}) \\
\textnormal{s.t.} \quad \mathbf{h}^t = h_\textrm{deep}(s(\mathbf{x}^{t-1}), \mathbf{h}^{t-1}, \mathbf{p}'\; | \;  \mathbf{W}) \quad \textnormal{and} \quad \mathbf{h}^0 = 0
\end{equation}

Notice the similarity between equations~\ref{eq:L} and \ref{eq:g}. Indeed, the equations for recurrent neural networks~(RNN) in equation~\ref{eq:g} are very similar to the ones of the loss of a physical model in equation~\ref{eq:L}. Therefore, we optimize this entire system as an RNN unfolded over time, as illustrated in Figure~\ref{fig:2}. The weights $\mathbf{W}$ are optimized with stochastic gradient descent. The gradient required for that is the Jacobian $d \mathcal{L} /d\mathbf{W} $, which is found with automatic differentiation software. 

We have now reduced the problem to a standard deep learning problem. We need to train our network $g_{\textrm{deep}}$ on a sufficient amount of samples $x^{\textrm{init}}$ and for a sufficient amount of sampled tasks $\mathbf{p}$ in order to get adequate generalization. Standard RNN regularization approaches could also improve this generalization. We reckon that generalization of $g_{\textrm{deep}}$ to more models $f_{\textrm{ph}}$, in order to ease the transfer of the controller from the model to the real system, is also possible~\citep{hermans2014automated}, but it is outside the scope of this paper.

\section{Results}

\subsection{Quadrupedal Robot: Computing Speed}
\begin{figure}[h!]
\begin{center}
\includegraphics[width=15cm]{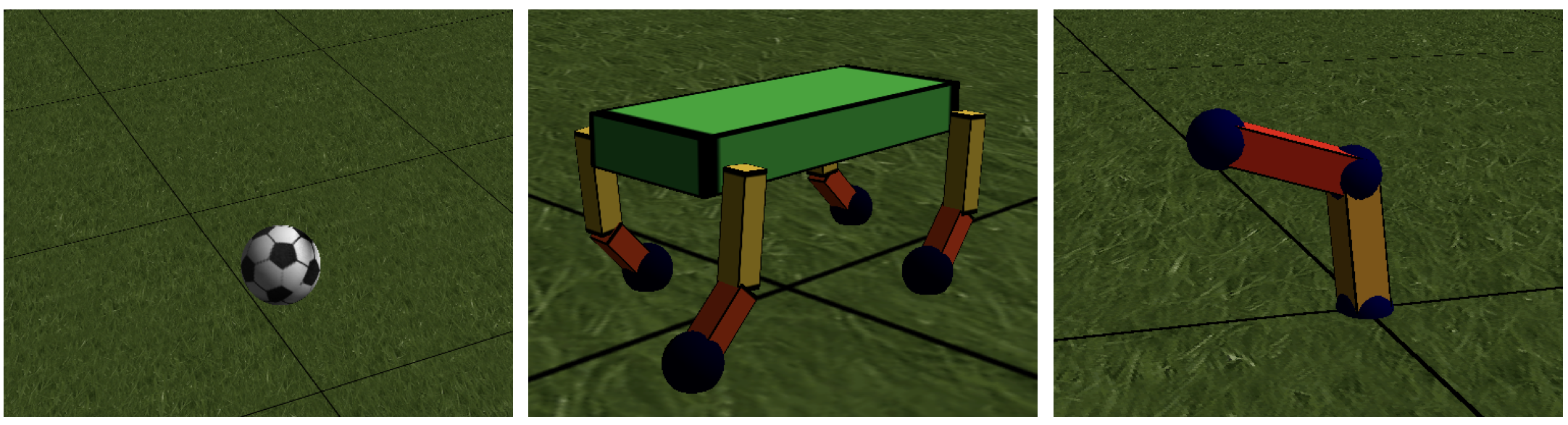}
\end{center}
\caption{\textbf{(A)} Illustration of the ball model used in the first task. \textbf{(B)} Illustration of the quadruped robot model with 8 actuated degrees of freedom, 1 in each shoulder, 1 in each elbow. The spine of the robot can collide with the ground, through 4 spheres in the inside of the cuboid. \textbf{(C)} Illustration of the robot arm model with 4 actuated degrees of freedom.}\label{fig:3}
\end{figure}

To verify the speed of our engine, we also implemented a small quadrupedal robot model, as illustrated in Figure~\ref{fig:3}\textbf{(B)}. This model has a total of 81 sensors, e.g. encoders and an inertial measurement unit (IMU). The servo motors are controlled in a closed loop by a small neural network $g_{\textrm{deep}}$ with a number of parameters, as shown in Figure~\ref{fig:2}. The gradient is the Jacobian of $\mathcal{L}$, the total traveled distance of the robot in \unit{10}{s} , differentiated with respect to all the parameters of the controller $\mathbf{W}$. This Jacobian is found by using BPTT and propagating all \unit{10}{s} back. The time it takes to compute this traveled distance and the accompanying Jacobian is shown in Table~\ref{speed}. We include both the computation time with and without the gradient, i.e. both the forward and backward pass and the forward pass alone. This way, the numbers can be compared to other physics engines, as those only calculate without gradient. Our implementation and our model can probably be made more efficient, and evaluating the gradient can probably be made faster a similar factor.

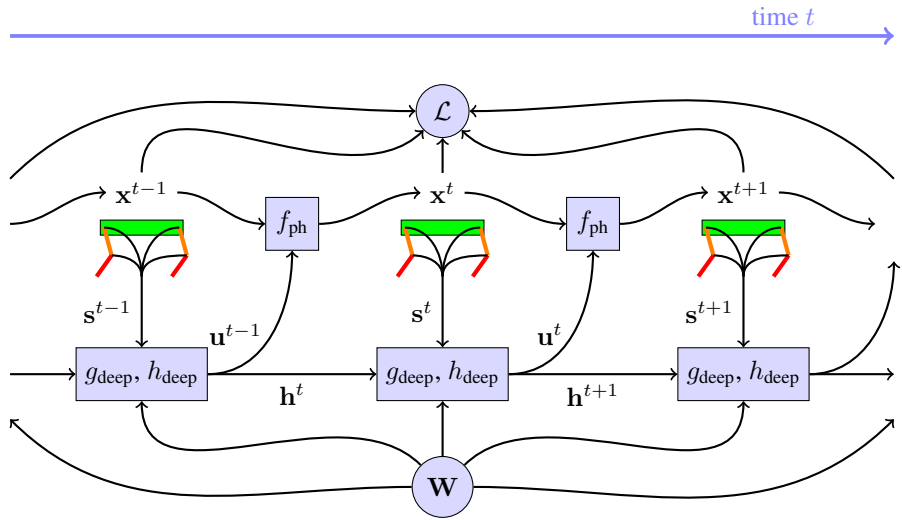
\begin{figure}[htpb]
\begin{center}
\begin{tikzpicture}  
    
    \draw[arrow,color=blue!50,line width=0.5mm] (-5.75, 2.5) to node [above,very near end] {time $t$} (6.0, 2.5);

    \path (-5.75, 0.6) coordinate (xhlt);
    \path (6.0, 0.6) coordinate (xhrt);
    
    \path (-5.75, -2.6) coordinate (ghhlb);
    \path (6.0, -2.6) coordinate (ghhrb);

    \path (-5.75, 0) coordinate (fphhl);
    
    \path (5.75, 0) coordinate (fphhr);
    \path (6.0, -0.5) coordinate (fphhrb);
    
    \path (-5.75, -2) coordinate (ghhl);
    \path (6.0, -2) coordinate (ghhr);

    \pic (rob0) at (-4,0) {robotsymbol};
    \node[anchor=west, above=0.1 of rob0-body] (x_t0) {$\mathbf{x}^{t-1}$};
    \draw[arrow,out=0,in=180] (fphhl) to (x_t0);
    
    \node[block] (gh0) at (-4,-2) {$g_{\textrm{deep}}$, $h_{\textrm{deep}}$};
    \draw[arrow] (ghhl) -- (gh0);
    \draw[arrow] (rob0-norm) -- node [left,midway] {$\mathbf{s}^{t-1}$} (gh0);
    
    \node[block] (fph0) at (-2,0) {$f_{\textrm{ph}}$};
    \draw[arrow,out=0,in=180] (x_t0) to (fph0);
    \draw[arrow,out=0,in=-90] (gh0) to node [left,midway] {$\mathbf{u}^{t-1}$} (fph0);

    \pic (rob1) at (0,0) {robotsymbol};
    \node[anchor=west, above=0.1 of rob1-body] (x_t1) {$\mathbf{x}^t$};
    \draw[arrow,out=0,in=180] (fph0) to (x_t1);
    
    \node[block] (gh1) at (0,-2) {$g_{\textrm{deep}}$, $h_{\textrm{deep}}$};
    \draw[arrow] (rob1-norm) -- node [left,midway] {$\mathbf{s}^{t}$} (gh1);    
    \draw[arrow] (gh0) -- node [below,midway] {$\mathbf{h}^t$} (gh1);
    
    \node[block] (fph1) at (2,0) {$f_{\textrm{ph}}$};
    \draw[arrow,out=0,in=180] (x_t1) to (fph1);
    \draw[arrow,out=0,in=-90] (gh1) to node [left,midway] {$\mathbf{u}^{t}$} (fph1);

    \pic (rob2) at (4,0) {robotsymbol};
    \node[anchor=west, above=0.1 of rob2-body] (x_t2) {$\mathbf{x}^{t+1}$};
    \draw[arrow,out=0,in=180] (fph1) to (x_t2);
    \draw[arrow,out=0,in=180] (x_t2) to (fphhr);
    
    \node[block] (gh2) at (4,-2) {$g_{\textrm{deep}}$, $h_{\textrm{deep}}$};
    \draw[arrow] (rob2-norm) -- node [left,midway] {$\mathbf{s}^{t+1}$} (gh2);
    \draw[arrow] (gh1) -- node [below,midway] {$\mathbf{h}^{t+1}$} (gh2);
    
    \draw[arrow,out=0,in=-90] (gh2) to (fphhrb);
    \draw[arrow] (gh2) -- (ghhr);

    \node[var] (W) at (0,-3.5) {$\mathbf{W}$};
    
    \draw[arrow,out=180,in=-45] (W) to (ghhlb);
    \draw[arrow,out=135,in=-90] (W) to (gh0);
    \draw[arrow,out=90,in=-90] (W) to (gh1);
    \draw[arrow,out=45,in=-90] (W) to (gh2);
    \draw[arrow,out=0,in=-135] (W) to (ghhrb);

    \node[var] (L) at (0,1.5) {$\mathcal{L}$};
    
    \draw[arrow,out=45,in=-180] (xhlt) to (L);
    \draw[arrow,out=90,in=-135] (x_t0) to (L);
    \draw[arrow,out=90,in=-90] (x_t1) to (L);
    \draw[arrow,out=90,in=-45] (x_t2) to (L);
    \draw[arrow,out=135,in=0] (xhrt) to (L);

\end{tikzpicture}
    
\end{center}
\caption{Illustration of the dynamic system with the robot and controller, after unrolling over time. The neural networks $g_{\textrm{deep}}$ and $h_{\textrm{deep}}$ with weights $\mathbf{W}$ receive sensor signals $\mathbf{s}^{t}$ from the sensors on the robot and use these to generate motor signals $\mathbf{u}^{t}$ which are used by the physics engine $f_{\textrm{ph}}$ to find the next state of the robot in the physical system. These neural networks also have a memory, implemented with recurrent connections $\mathbf{h}^{t}$. From the state $\mathbf{x}^t$ of these robots, the loss $\mathcal{L}$ can be found. In order to find $d \mathcal{L} /d\mathbf{W} $, every block in this chart needs to be differentiable. The contribution of this paper, is to implement a differentiable $f_{\textrm{ph}}$, which allows us to optimize $\mathbf{W}$ to minimize $\mathcal{L}$ more efficiently than was possible before. } \label{fig:2}
\end{figure}

When only a single controller is optimized, our engine runs more slowly on GPU than on CPU. To tackle this issue, we implemented batch gradient descent, which is commonly used in complex optimization problems. In this case, by batching our robot models, we achieve significant acceleration on GPU. Although backpropagating the gradient through physics slows down the computations by roughly a factor 10, this factor only barely increases with the number of parameters in our controller. 

Combining this with our previous observation that fewer iterations are needed when using gradient descent, our approach can enable the use of gradient descent through physics for highly complex deep neural network controllers with millions of parameters. Also note that by using a batch method, a single GPU can simulate about 864 000 model seconds per day, or 86 400 000 model states. This should be plenty for deep learning. It also means that a single simulation step of a single robot, which includes collision detection, solving the LCP problem, integrating the velocities and backpropagating the gradient through it all, takes about \unit{1}{ms} on average. Without the backpropagation, this process is only about seven times faster.

\subsection{4 Degree of Freedom Robot Arm}
\label{tricks}
As a first test of optimizing robot controllers, we implemented a four degree of freedom robotic arm, as depicted in Figure~\ref{fig:3}\textbf{(C)}. The bottom of the robot has a 2 degrees of freedom actuated universal joint; the elbow has a 2 degree of freedom actuated joint as well. The arm is \unit{1}{m} long, and has a total mass of \unit{32}{kg}. The servos have a gain of \unit{30}{s^{-1}}, a torque of \unit{30}{N m} and a velocity of \unit{45\degree}{s^{-1}}.

For this robot arm, we train controllers for a task with a gradually increasing amount of difficulty. To be able to train our parameters, we have to use a couple of tricks often used in the training of recurrent neural networks. 
\begin{itemize}
    \item We choose an objective which is evaluated at every time step and then averaged, rather than at specific points of the simulation. This  approach vastly increases the number of samples over which the gradient is averaged, which in turn makes the gradient direction more reliable~\citep{sjoberg1995nonlinear}.
    \item The value of the gradient is decreased by a factor $\alpha < 1$ at every time step. This trick has the effect of a prior. Namely, events further in the past are less important for influencing current events, because intermediate events might diminish their influence altogether. It also improves robustness against exploding gradients~\citep{hermans2014automated}.
    \item We initialize the controller intelligently. We do not want the controller to shake the actuators violently and explore outside the accurate domain of our simulation model. Therefore our controllers are initialized with zeros such that they only output zeros at the start of the simulation. The initial policy is the zero policy.
    \item We constraint the size of the gradient to an L2-norm of 1. This makes sure that gradients close to discontinuities in the fitness landscape do not push the parameter values too far away, such that everything which was learned is forgotten~\citep{sutskever2013training}.
\end{itemize}

\subsubsection{Reaching a Fixed Point}

A first simple task, is to have a small neural net controller learn to move the controller to a certain fixed point in space, at coordinates $(\unit{0.5}{m};\unit{0.5}{m};\unit{0.5}{m})$. The objective we minimize for this task, is the distance between the end effector and the target point, averaged over the 8 seconds we simulate our model. 

We provide the controller with a single sensor input, namely the current distance between the end effector and the target point. Input is not required for this task, as there are solutions for which the motor signals are constant in time. However, this would not necessarily be the optimal approach for minimizing the average distance over time, it only solves the distance at the end of the simulation, but does not minimize the distance during the trajectory to get at the final position.

As a controller, we use a dense neural network with 1 input, 2 hidden layers of 128 units with a rectifier activation function, and 4 outputs with an identity activation function. Each unit in the neural network also has a bias parameter. This controller has 17 284 parameters in total. We disabled the recurrent connections $\mathbf{h}^{t}$.

We use gradient descent with a batch size of 1 robot for optimization, as the problem is not stochastic in nature. The parameters are optimized with Adam's rule~\citep{kingma2014} with a learning rate of 0.001. Every update step with this method takes about 5 seconds on CPU. We find that the controller comes within \unit{4}{cm} of the target in 100 model evaluations, and within \unit{1}{cm} in 150 model evaluations, which is small compared to the \unit{1}{m} arm of the robot. Moreover, the controller does find a more optimal trajectory which takes into account the sensor information.

Solving problems like these in fewer iteration steps than the number of parameters, is unfeasible with derivative free methods~\citep{sjoberg1995nonlinear}. Despite that, we did try to optimize the same problem with CMA-ES. After a week of computing and 60 000 model evaluations, CMA-ES did not show any sign of improvement nor convergence, as it cannot handle the sheer amount of parameters. In performance, the policy went from a starting performance of $\unit{0.995\pm0.330}{m}$ to a not significantly different $\unit{0.933\pm0.369}{m}$ after the optimization. For this reason, we did not continue using CMA-ES as a benchmark in the further experiments.

\subsubsection{Reaching a Random Point}

As a second task, we sample a random target point in the reachable space of the end effector. We give this point as input $v'$ to the controller, and the task is to again minimize the average distance between the end effector and the target point $v$. Our objective $\mathcal{L}$ is this distance averaged over all timesteps.

As a controller, we use a dense neural network comparable to the previous section, but this time with 3 inputs. Note that this is an open loop controller, which needs to control the system to a set point given as input. We used 3 hidden layers with 1024 units each, so the controller has  2~ 107~396 parameters in total. This is not necessary for this task, but we do it like this to demonstrate the power of this approach. In order to train for this task, we use a batch size of 128 robots, such that every update step takes \unit{58}{s} on GPU. Each simulation takes \unit{8}{s} with a simulation step of \unit{0.01}{s}. Therefore, the gradient on the parameters of the controllers has been averaged over 51 200 timesteps at every update step. We update the parameters with Adam's rule, where we scale the learning rate with the average error achieved in the previous step.

We find that it takes 576 update steps before the millions of parameters are optimized, such that the end effector of the robot is on average less than \unit{10}{cm} of target, 2 563 update steps before the error is less than \unit{5}{cm}. 






\subsection{A Quadrupedal Robot: revisited}

Optimizing a gait for a quadrupedal robot is a problem of a different order, something the authors have extensive experience with~\citep{sproewitz2013towards,degrave2013comparing,degrave2015transfer}. The problem is way more challenging and allows for a broad range of possible solutions. In nature, we find a wide variety of gaits, from hopping over trotting, walking and galloping. With hand tuning on the robot model shown in Figure~\ref{fig:3}\textbf{(B)}, we were able to obtain a trotting motion with an average forward speed of \unit{0.7}{m/s}. We found it tricky to find a gait where the robot did not end up like an upside down turtle, as 75\% of the mass of the robot is located in its torso.

As a controller for our quadrupedal robot, we use a neural network with 2 input signals $\mathbf{s}^t$, namely a sine and a cosine signal with a frequency of \unit{1.5}{Hz}. On top of this, we added 2 hidden layers of 128 units and a rectifier activation function. As output layer, we have a dense layer with 8 units and a linear activation function, which has as input both the input layer and the top layer of the hidden layers. In total, this controller has 17 952 parameters. Since the problem is not stochastic in nature, we use a batch size of 1 robot. We initialize the output layer with zero weights, so the robot starts the optimization in a stand still position.

We optimize these parameters to maximize the average velocity of the spine over the course of \unit{10}{s} of time in simulation. This way, the gradient used in the update step is effectively an average of the 1 000 time steps after unrolling the recurrent connections. This objective does not take into account energy use, or other metrics typically employed in robotic problems.

In only 500 model evaluations or about 1 hour of optimizing on CPU, the optimization with BPTT comes up with a solution with a speed of \unit{1.17}{m/s}. This solution is a hopping gait, with a summersault every 3 steps\footnote{A video is available at \url{https://goo.gl/5ykZZe}}, despite limiting the torque of the servos to \unit{4}{Nm} on this \unit{28.7}{kg} robot. For more life-like gaits, energy efficiency could be use as a regularization method. Evaluating these improvements are however outside the scope of this paper.


\subsection{The inverted pendulum with a camera as sensor}

As a fourth example, we implemented a model of the pendulum-cart system we have in our laboratorium. This pendulum-cart system is used for the classic control task of the underactuated inverted pendulum~\citep{vaccaro1995digital}. In this example however, a camera which is set up in front of the system is the only available information for the controller. It therefore has to observe the system it controls using vision, i.e., learning from pixels. A frame captured by this camera is shown in Figure~\ref{fig:4}.

\begin{figure}[htpb]
    \begin{center}
        \includegraphics[width=0.5\columnwidth]{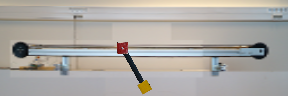}
        \caption{A frame captured by the differentiable camera looking at the model of the pendulum-cart system. The resolution used is 288 by 96 pixels. All the textures are made from pictures of the actual system.}
        \label{fig:pinhole}
    \end{center}
\end{figure}

\begin{figure}[tpb]
\begin{center}

\begin{tikzpicture}
\usetikzlibrary{calc}

\def\xOne{1}
\def\xTwo{0.5}
\def\yOne{0}
\def\yTwo{-1.3}
\def\zOne{-1}
\def\zTwo{0.5}

\draw[very thick,->] (-\zOne/2,-\zTwo/2) -- (\zOne/2,\zTwo/2) node[anchor=north,yshift=-2pt,xshift=3pt,font=\footnotesize]{$d$};
\node at (0.75,-0.5) {$\mathcal{C}$};

\draw[very thin,solid] (-\zOne/2-2*\xOne,-\zTwo/2-2*\xTwo) -- (-\zOne/2+2*\xOne,-\zTwo/2+2*\xTwo); 
\draw[very thin,solid] (3*\zOne,3*\zTwo) -- (6*\zOne,6*\zTwo); 

\draw[very thin,solid] (6*\zOne-2*\xOne,6*\zTwo-2*\xTwo) -- (6*\zOne+2*\xOne,6*\zTwo+2*\xTwo) node[anchor=west]{}; 

\draw[-latex,line width=3pt,blue,line cap=round] (6*\zOne+1.4*\xOne,6*\zTwo+1.4*\xTwo) -- (6*\zOne+1.4*\xOne,6*\zTwo+1.4*\xTwo+1.1) node[anchor=south,font=\footnotesize]{$ P = (X,Y,Z) $};
\node[circle,inner sep=0pt,minimum size=0.2cm,fill=blue] (object) at (6*\zOne+1.4*\xOne,6*\zTwo+1.4*\xTwo+1.1) {};

\draw[thick,solid,red] (3*\zOne+0.69*\xOne,3*\zTwo+0.7*\xTwo+0.69) -- (6*\zOne+1.4*\xOne,6*\zTwo+1.4*\xTwo+1.1);

\filldraw[fill=gray!20,draw=gray!70,opacity=0.8] (3*\zOne-1.5*\xOne-1.5*\yOne,3*\zTwo-1.5*\xTwo-1.5*\yTwo) -- (3*\zOne+1.5*\xOne-1.5*\yOne,3*\zTwo+1.5*\xTwo-1.5*\yTwo) -- (3*\zOne+1.5*\xOne+1.5*\yOne,3*\zTwo+1.5*\xTwo+1.5*\yTwo) -- (3*\zOne-1.5*\xOne+1.5*\yOne,3*\zTwo-1.5*\xTwo+1.5*\yTwo) -- (3*\zOne-1.5*\xOne-1.5*\yOne,3*\zTwo-1.5*\xTwo-1.5*\yTwo);

\draw[->,thick,green!70!black,dashed] (3*\zOne-1.5*\xOne-1.5*\yOne,3*\zTwo-1.5*\xTwo-1.5*\yTwo) -- (3*\zOne+2*\xOne-1.5*\yOne,3*\zTwo+2*\xTwo-1.5*\yTwo)
     node[anchor=north west, xshift=-3pt,font=\footnotesize]{};
\draw[->,thick,green!70!black,dashed] (3*\zOne-1.5*\xOne-1.5*\yOne,3*\zTwo-1.5*\xTwo-1.5*\yTwo) -- (3*\zOne-1.5*\xOne-1.5*\yOne,3*\zTwo-1.5*\xTwo+2*\yTwo)
     node[anchor=west,font=\footnotesize]{};

\draw[-,thick,cyan,dashed] (3*\zOne-2*\xOne,3*\zTwo-2*\xTwo) -- (3*\zOne+2*\xOne,3*\zTwo+2*\xTwo)
     node[anchor=north west, xshift=-3pt,font=\footnotesize]{};
\draw[-,thick,cyan,dashed] (3*\zOne-2*\yOne,3*\zTwo-2*\yTwo) -- (3*\zOne+2*\yOne,3*\zTwo+2*\yTwo)
     node[anchor=west,font=\footnotesize]{};

\draw[-latex,line width=1.5pt,blue,line cap=round] (3*\zOne+0.69*\xOne,3*\zTwo+0.69*\xTwo) -- (3*\zOne+0.69*\xOne,3*\zTwo+0.69*\xTwo+0.69);
\node[circle,inner sep=0pt,minimum size=0.1cm,fill=blue] (object) at (3*\zOne+0.69*\xOne,3*\zTwo+0.7*\xTwo+0.69) {};

\filldraw[red,opacity=0.6] (3*\zOne+6*0.105*\xOne,3*\zTwo+0.75+6*0.105*\xTwo) -- ++(0.105*\xOne,0.105*\xTwo) -- ++(0.105*\yOne,0.105*\yTwo) -- ++(-0.105*\xOne,-0.105*\xTwo) -- ++(-0.105*\yOne,-0.105*\yTwo);

\draw[thick,solid,red] (-\zOne/2,-\zTwo/2) -- (3*\zOne+0.69*\xOne,3*\zTwo+0.7*\xTwo+0.69);

\draw[thin,solid] (0,0) -- (3*\zOne,3*\zTwo);

\draw (3*\zOne-1*\xOne+1.3*\yOne,3*\zTwo-1*\xTwo+1.3*\yTwo) node[gray!70,rotate=28] {$ z = f $};
\node[green!70!black,anchor=west,font=\scriptsize] at (3*\zOne+0.69*\xOne,3*\zTwo+0.7*\xTwo+0.69) {$ (u,v) $};
\draw[very thin] (5.5*\zOne-0.02*\xOne+0.02*\yOne,5.5*\zTwo-0.02*\xOne+0.02*\yOne) .. controls (5.5*\zOne-0.1*\xOne+0.3*\yOne,5.5*\zTwo-0.1*\xTwo+0.3*\yTwo) and (5.5*\zOne-0.3*\xOne+0.1*\yOne,5.5*\zTwo-0.3*\xTwo+0.1*\yTwo) ..  (5.5*\zOne-0.6*\xOne+0.4*\yOne,5.5*\zTwo-0.6*\xTwo+0.4*\yTwo) node[anchor=north,align=center,font=\sffamily\scriptsize] {optical \\ axis};

\draw[thin,gray!70] (3*\zOne,3*\zTwo+0.75) -- ++(0.105*\xOne,0.105*\xTwo) -- ++(0.105*\yOne,0.105*\yTwo) -- ++(-0.105*\xOne,-0.105*\xTwo) -- ++(-0.105*\yOne,-0.105*\yTwo) -- ++(0.21*\xOne,0.21*\xTwo) -- ++(0.105*\yOne,0.105*\yTwo) -- ++(-0.105*\xOne,-0.105*\xTwo) -- ++(-0.105*\yOne,-0.105*\yTwo) -- ++(0.21*\xOne,0.21*\xTwo) -- ++(0.105*\yOne,0.105*\yTwo) -- ++(-0.105*\xOne,-0.105*\xTwo) -- ++(-0.105*\yOne,-0.105*\yTwo) -- ++(0.21*\xOne,0.21*\xTwo) -- ++(0.105*\yOne,0.105*\yTwo) -- ++(-0.105*\xOne,-0.105*\xTwo) -- ++(-0.105*\yOne,-0.105*\yTwo) -- ++(0.21*\xOne,0.21*\xTwo) -- ++(0.105*\yOne,0.105*\yTwo) -- ++(-0.105*\xOne,-0.105*\xTwo) -- ++(-0.105*\yOne,-0.105*\yTwo) -- ++(0.21*\xOne,0.21*\xTwo) -- ++(0.105*\yOne,0.105*\yTwo) -- ++(-0.105*\xOne,-0.105*\xTwo) -- ++(-0.105*\yOne,-0.105*\yTwo) -- ++(0.21*\xOne,0.21*\xTwo) -- ++(0.105*\yOne,0.105*\yTwo) -- ++(-0.105*\xOne,-0.105*\xTwo) -- ++(-0.21*\yOne,-0.21*\yTwo) -- ++(0.105*\xOne,0.105*\xTwo) -- ++(0.105*\yOne,0.105*\yTwo) -- ++(-0.105*\xOne,-0.105*\xTwo) -- ++(-0.21*\yOne,-0.21*\yTwo) -- ++(0.105*\xOne,0.105*\xTwo) -- ++(0.105*\yOne,0.105*\yTwo) -- ++(-0.105*\xOne,-0.105*\xTwo) -- ++(-0.21*\yOne,-0.21*\yTwo) -- ++(0.105*\xOne,0.105*\xTwo) -- ++(0.105*\yOne,0.105*\yTwo) -- ++(-0.105*\xOne,-0.105*\xTwo) -- ++(-0.21*\yOne,-0.21*\yTwo) -- ++(0.105*\xOne,0.105*\xTwo) -- ++(0.105*\yOne,0.105*\yTwo) -- ++(-0.105*\xOne,-0.105*\xTwo) -- ++(-0.21*\yOne,-0.21*\yTwo) -- ++(0.105*\xOne,0.105*\xTwo) -- ++(0.105*\yOne,0.105*\yTwo) -- ++(-0.105*\xOne,-0.105*\xTwo) -- ++(-0.21*\yOne,-0.21*\yTwo) -- ++(0.105*\xOne,0.105*\xTwo) -- ++(0.105*\yOne,0.105*\yTwo) -- ++(-0.105*\xOne,-0.105*\xTwo) -- ++(-0.21*\yOne,-0.21*\yTwo) -- ++(0.105*\xOne,0.105*\xTwo) -- ++(0.105*\yOne,0.105*\yTwo) -- ++(-0.105*\xOne,-0.105*\xTwo) -- ++(-0.21*\yOne,-0.21*\yTwo) -- ++(0.105*\xOne,0.105*\xTwo) -- ++(0.105*\yOne,0.105*\yTwo) -- ++(-0.105*\xOne,-0.105*\xTwo) -- ++(-0.19*\yOne,-0.19*\yTwo) -- ++(0.105*\xOne,0.105*\xTwo) -- ++(0.105*\yOne,0.105*\yTwo);

\end{tikzpicture}
\end{center}
\caption{The camera model used to convert the three dimensional point $P$ into a two dimensional pixel on the projection plane $(u,v)$.} \label{fig:2}
\end{figure}
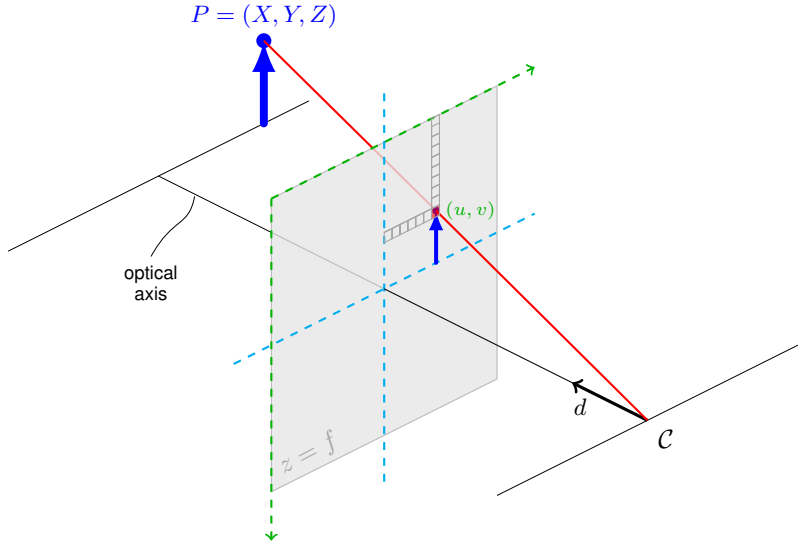

In order to build this model, we implemented a renderer in our physics engine which converts the three dimensional scene into a two dimensional color image, as illustrated in Figure~\ref{fig:pinhole}. In order to perform this operation in a differentiable way, we use a ray tracing approach rather than the more conventional rasterization pipeline. First we cast a set of lines from the point of our camera $\mathcal{C}$ in the direction $\vec{d}$ of the optical axis of the camera. These vectors are then converted with the pinhole camera model into a line going through the center of the pixel with the image coordinates $(u, v)$ on the projection plane. Each of these rays is then intersected with every object in the scene to find the texture and corresponding sample location to sample from in the scene's texture array. From all intersections a single ray makes, all but the one closest in front of the projection plane is kept.

Each of the intersections is then converted to a color by bilinearly interpolating the scene's texture array, in a way similar to the approach used for the spatial transform layer~\citep{jaderberg2015spatial, degrave2016spatial}. This bilinear interpolation is necessary to make the frame captured by the camera differentiable to the state of the robot with non-zero derivatives. If the textures would have been a zero-order, pixelated approximation, then all the gradients would be zero analytically. 

Using the above ray-tracing approach, we minimize the distance from the end of the pendulum to the desired point and regularize the speed of the pendulum. The memoryless deep controller receives the current image of the camera, in addition to two images from the past such that it can estimate velocity and acceleration. We observe that a controller with 1,065,888 parameters is able to learn to swing up and keep the pendulum stable after only 2420 episodes of 3 model seconds. The complete optimization process took 15 hours on 1 GPU. The resulting controller keeps the pendulum stable for more than one minute\footnote{https://twitter.com/317070/status/821062814798331905}. In order to do this, the controller has learned to interpret the frames it receives from the camera and found a suitable control strategy.

Note that this would not have been possible using a physics engine such as mujoco, as these engines only allow differentiation through the action and the state, but does not allow to differentiate through the renderer. We want to stress that in this setup we solved the problem by backpropagating through both the computer vision in the form of the convolutional neural network, and the renderer in the form of the differentiable camera.

\section{Discussion}
We implemented a modern engine which can run a 3D rigid body model, using the same algorithm as other engines commonly used to simulate robots, but we can additionally differentiate control parameters with BPTT. Our implementation also runs on GPU, and we show that using GPUs to simulate the physics can speed up the process for large batches of robots. We show that even complex sensors such as cameras, can be implemented and differentiated through, allowing for computer vision to be learned together with a control policy.

When initially addressing the problem, we did not know whether finding the gradient would be computationally tractable, let alone whether evaluating it would be fast enough to be beneficial for optimization. In this paper, we have demonstrated that evaluating the gradient is tractable enough to speed up optimization on problems with as little as six parameters. The speed of this evaluation mainly depends on the complexity of the physics model and only slightly on the number of parameters to optimize. Therefore, our results suggest that this cost is dominated by the gain achieved from the combination of using batch gradient descent and GPU acceleration. Consequently, by using gradient descent with BPTT one can speed up the optimization processes often found in robotics, even for rather small problems, due to the reduced number of model evaluations required. Furthermore, this this improvement in speed scales to problems with a lot of parameters. By using the proposed engine, finding policies for robot models can be done faster and in a more straightforward way. This method should allow for a new approach to apply deep learning techniques in robotics.

Optimizing the controller of a robot model with gradient-based optimization is equivalent to optimizing an RNN. After all, the gradient passes through each parameter at every time step. The parameter space is therefore very noisy. Consequently, training the parameters of this controller is a highly non-trivial problem, as it corresponds to training the parameters of an RNN. On top of that, exploding and vanishing signals and gradients cause far more challenging problems compared to feed forward networks. 

In section~\ref{tricks}, we already discussed some of the tricks used for optimizing RNNs. Earlier research shows that these methods can be extended to more complicated tasks than the ones discussed here~\citep{hermans2014automated,sutskever2013training}. Hence, we believe that this approach towards learning controllers for robotics applies to more complex problems than the illustrative examples in this paper.


All of the results in this paper will largely depend on showing how these controllers will work on the physical counterparts of our models. Nonetheless, we would like to conjecture that to a certain extent, this gradient of a model is close to the gradient of the physical system. The gradient of the model is more susceptible to high-frequency noise introduced by modeling the system, than the imaginary gradient of the system itself. Nonetheless, it contains information which might be indicative, even if it is not perfect. We would theorize that using this noisy gradient is still better than optimizing in the blind and that the transferability to real robots can be improved by evaluating the gradients on batches of (slightly) different robots in (slightly) different situations and averaging the results. This technique has already been applied in \citep{hermans2014automated} as a regularization method to avoid bifurcations during online learning. If the previous proves to be correct, our approach can offer an addition or possibly even an alternative to deep Q-learning for deep neural network controllers in robotics.

We can see the use of this extended approach for a broad range of applications in robotics. Not only do we think there are multiple ways where recent advances in deep learning could be applied to robotics more efficiently with a differentiable physics engine, we also see various ways in which this engine could improve existing angles at which robotics are currently approached:

\begin{itemize}

    \item In this paper, we added memory by introducing recurrent connections in the neural network controller. We reckon that advanced, recurrent connections such as ones with a memory made out of LSTM cells~\citep{hochreiter1997long} can allow for more powerful controllers than the controllers described in this paper.
    \item Having general differentiable models should allow for an efficient system identification process~\citep{bongard2006resilient,ha2018world}. The physics engine can find analytic derivatives to all model parameters. This includes masses and lengths, but also parameters which are not typically touched in system identification, such as the textures of the rigid body. As the approach could efficiently optimize many parameters simultaneously, it would be conceivable to find state dependent model parameters using a neural network to map the current state onto e.g. the friction coefficient in that state.
    \item Using a differentiable physics engine, we reckon that knowledge of a model can be distilled more efficiently into a forward or backward model in the form of a neural network, similar to methods such as used in~\citet{johnson2016perceptual} and \citet{dumoulin}. By differentiating through an exact model and defining a relevant error on this model, it should be possible to transfer knowledge from a forward or backward model in the differentiable physics engine to a forward or backward neural network model. Neural network models trained this way might be more robust than the ones learned from generated trajectories~\citep{christiano2016transfer}. In turn, this neural model could then be used for faster but approximate evaluation of the model.
    \item Although we did not address this in this paper, there is no reason why only control parameters could be optimized in the process. Hardware parameters of the robot have been optimized the same way before~\citep{jarny1991general,iollo2001aerodynamic,hermans2014automated}. The authors reckon that the reverse process is also true. A physics engine can provide a strong prior, which can be used for robots to learn (or adjust) their robot models based on their hardware measurements faster than today. You could optimize the model parameters with gradient descent through physics, to have the model better mimic the actual observations.
    \item Where adversarial networks are already showing their use in generating image models, we believe adversarial robotics training (ART) will create some inventive ways to design and control robots. Like in generative adversarial nets (GAN)~\citep{goodfellow2014generative}, where the gradient is pulled through two competing neural networks, the gradient could be pulled through multiple competing robots as well. It would form an interesting approach for swarm robotics, similar to previous results in evolutionary robotics~\citep{sims1994evolving,pfeifer2006body,cheney2014evolved}, but possibly faster.
\end{itemize}    

\section*{Tables}

\begin{table}[h]
    \renewcommand{\arraystretch}{1.3}
    \centering
    \caption{Evaluation of the computing speed of our engine on a robot model controlled by a closed loop controller with a variable number of parameters. We evaluated both on  CPU (i7 5930K) and GPU (GTX 1080), both for a single robot optimization and for batches of multiple robots in parallel. The numbers are the time required in seconds for simulating the quadruped robot(s) for \unit{10}{s}, with and without updating the controller parameters through gradient descent. The gradient calculated here is the Jacobian of the total traveled distance of the robot in \unit{10}{s}, differentiated with respect to all the parameters of the controller. For comparison, the model has 102 states. It is built from 17 rigid bodies, each having 6 degrees of freedom. These states are constrained by exactly 100 constraints.}
    \setminimalcolor{0}
    \setmaximalcolor{311}
	\begin{tabular}{rr*{1}{cc@{\hskip 0.3in}}cc}
		\multicolumn{6}{c}{\emph{Seconds of computing time required to simulate a batch of robots for 10 seconds}}\\
	    & & \multicolumn{2}{c@{\hskip 0.3in}}{with gradient} & \multicolumn{2}{c@{\hskip 0.15in}}{without gradient} \\
		& & CPU & GPU & CPU & GPU \\
		\multirow{ 2}{*}{1 robot} & 1\,296 parameters & \gyr{8.17} & \gyr{69.6} & \gyr{1.06} & \gyr{9.69} \\\vspace{0.15in}
		& 1\,147\,904 parameters & \gyr{13.2} & \gyr{75.0} & \gyr{2.04} & \gyr{9.69} \\
		\multirow{ 2}{*}{128 robots} & 1\,296 parameters & \gyr{263} & \gyr{128} & \gyr{47.7} & \gyr{17.8} \\
		& 1\,147\,904 parameters & \gyr{311} & \gyr{129} & \gyr{50.4} & \gyr{18.3} \\
	\end{tabular}
	\vspace{3em}
    \setminimalcolor{0}
    \setmaximalcolor{75}
	\begin{tabular}{rr*{1}{cc@{\hskip 0.3in}}cc}
	    \\
	    \\
		\multicolumn{6}{c}{\emph{Milliseconds of computing time required to perform one time step of one robot.}}\\
	    & & \multicolumn{2}{c@{\hskip 0.3in}}{with gradient} & \multicolumn{2}{c@{\hskip 0.15in}}{without gradient} \\
		& & CPU & GPU & CPU & GPU \\
		\multirow{ 2}{*}{1 robot} & 1\,296 parameters & \gyr{8.17} & \gyr{69.6} & \gyr{1.06} & \gyr{9.69} \\\vspace{0.15in}
		& 1\,147\,904 parameters & \gyr{13.2} & \gyr{75.0} & \gyr{2.04} & \gyr{9.69} \\
		\multirow{ 2}{*}{128 robots} & 1\,296 parameters & \gyr{2.05} & \gyr{1.00} & \gyr{0.372} & \gyr{0.139} \\
		& 1\,147\,904 parameters & \gyr{2.43} & \gyr{1.01} & \gyr{0.394} & \gyr{0.143} \\
	\end{tabular}
    \label{speed}
\end{table}

\section*{Conflict of Interest Statement}
The authors declare that the research was conducted in the absence of any commercial or financial relationships that could be construed as a potential conflict of interest.

\section*{Author Contributions}

The experiments were conceived by dr. J. Degrave, dr. M. Hermans, prof. J. Dambre and prof. F. wyffels and designed by dr. J. Degrave and dr. M. Hermans. The data were analyzed by dr. J. Degrave with help of prof. F. wyffels and prof. J. Dambre. The manuscript was mostly written by dr. J. Degrave, with comments and corrections from prof. F. wyffels and prof. J. Dambre.

\section*{Funding}
The research leading to these results has received funding from the Agency for Innovation by Science and Technology in Flanders (IWT). The NVIDIA Corporation donated the GTX 1080 used for this research.

\section*{Acknowledgments}
Special thanks to David Pfau for pointing out relevant prior art we were previously unaware of, and Iryna Korshunova for proofreading the paper. 



\bibliographystyle{frontiersinSCNS_ENG_HUMS} 
\bibliography{frontiersin}

\end{document}